# Online Semi-Supervised Learning on Quantized Graphs

**Michal Valko**
Computer Science Department
University of Pittsburgh

**Branislav Kveton**
Intel Labs
Santa Clara, CA

**Ling Huang**
Intel Labs
Berkeley, CA

**Daniel Ting**
Department of Statistics
University of California Berkeley

## Abstract

In this paper, we tackle the problem of online semi-supervised learning (SSL). When data arrive in a stream, the dual problems of computation and data storage arise for any SSL method. We propose a fast approximate online SSL algorithm that solves for the harmonic solution on an approximate graph. We show, both empirically and theoretically, that good behavior can be achieved by collapsing nearby points into a set of local "representative points" that minimize distortion. Moreover, we regularize the harmonic solution to achieve better stability properties. We apply our algorithm to face recognition and optical character recognition applications to show that we can take advantage of the manifold structure to outperform the previous methods. Unlike previous heuristic approaches, we show that our method yields provable performance bounds.

## 1 INTRODUCTION

Semi-supervised learning (SSL) is a field of machine learning that studies learning from both labeled and unlabeled examples. This learning paradigm is extremely useful for solving real-world problems, where data is often abundant but the resources to label them are limited. Many SSL algorithms have been recently proposed (Zhu, 2008). One popular method is to compute the harmonic solution (HS) on the similarity graph (Zhu, Ghahramani, & Lafferty, 2003), and use it to infer labels of unlabeled examples.

In this paper we investigate an online learning formulation of SSL, which is suitable for *adaptive* machine learning systems. In this setting, a few labeled examples are provided in advance and set the initial bias of the system while unlabeled examples are gathered online and update the bias continuously. In the online setting, learning is viewed as a repeated game against a potentially adversarial nature. At each step $t$ of this game, we observe an example $\mathbf{x}_t$, and then predict its label $\hat{y}_t$. The challenge of the game is that after the game started we do not observe the true label $y_t$. Thus, if we want to adapt to changes in the environment, we have to rely on indirect forms of feedback, such as the structure of data.

Despite the usefulness of this paradigm for practical adaptive algorithms (Grabner, Leistner, & Bischof, 2008; Goldberg, Li, & Zhu, 2008), there is not much success in applying this paradigm to realistic problems, especially when data arrive at a high rate such as in video applications. Grabner et al. (2008) applies online semi-supervised boosting to object tracking, but uses a heuristic method to greedily label the unlabeled examples. This method learns a binary classifier, where one of the classes explicitly models outliers. In comparison, our approach is multi-class and allows for implicit modeling of outliers. The two algorithms are compared empirically in Section 5. Goldberg et al. (2008) develop an online version of manifold regularization of SVMs. Their method learns max-margin decision boundaries, which are additionally regularized by the manifold. Unfortunately, the approach was never applied to a naturally online learning problem, such as adaptive face recognition. Moreover, while the method is sound in principle, no theoretical guarantees are provided.

In this paper we focus on developing a fast approximate algorithm and prove performance guarantees for online SSL. Given $n_u$ unlabeled data points and a typical graph construction method, exact computation of HS has a space and time complexity of $\Omega(n_u^2)$ in general due to the construction of an $n_u \times n_u$ similarity matrix. Furthermore, exact computation requires an inverse operation on an $n_u \times n_u$ similarity matrix which takes $O(n_u^3)$ in most practical implementations[1]. For applications with large data size (e.g., exceeding thousands), the exact computation or even storage of HS becomes infeasible.

---
[1] The complexity can be further improved to $\mathcal{O}(n_u^{2.376})$ by using the Coppersmith-Winograd algorithm.

An influential line of work in the related area of graph partitioning approaches the computation problem by reducing the size of the graph, collapsing vertices and edges, partitioning the smaller graph, and then uncoarsening to construct a partition for the original graph (Hendrickson & Leland, 1995; Karypis & Kumar, 1999). Our work is similar in spirit but provides a theoretical analysis for a particular kind of coarsening and uncoarsening methodology.

Our aim is to find an effective *data preprocessing* technique that reduces the size of the data and coarsens the graph (Madigan, Raghavan, Dumouchel, Nason, Posse, & Ridgeway, 2002; Mitra, Murthy, & Pal, 2002). Some widely used data preprocessing approaches are based on data quantization, which replaces the original data set with a small number of centroids that capture relevant structure (Goldberg et al., 2008; Yan, Huang, & Jordan, 2009). Such approaches are often heuristic and do not quantify the relationship between the noise induced by the quantization and the final prediction risk.

An alternative approach to the computation problem is the *Nyström method*, a low rank matrix approximation method that allows faster computation of the inverse. This method has been widely adopted, particularly in the context of approximations for SVMs (Drineas & Mahoney, 2005; Williams & Seeger, 2001; Fine & Scheinberg, 2001) and spectral clustering (Fowlkes, Belongie, Chung, & Malik, 2004). However, since the Nyström method uses interactions between subsampled points and *all* other data points, storage of all points is required and thus, it becomes unsuitable for infinitely streamed data. To our best knowledge, we are not aware of any online version of Nyström method that could process an unbounded amount of streamed data. In addition, in an offline setting, Nyström-based methods have inferior performance than the quantization-based methods, if both of them are given the same time budget for computation, which was shown in an early work on the spectral clustering (Yan et al., 2009).

We propose an online SSL algorithm that uses a combination of quantization and regularization to obtain fast and accurate performance with provable performance bounds. Our algorithm takes the following form: 1) coarsen the similarity graph by replacing the neighboring points with a set of centroids, 2) solve for a regularized HS on the reduced graph, and 3) predict labels on the original data points based on predictions on the centroids. Extending the solution from the reduced graph to an approximate solution on the full graph is fast, taking time $O(n_u)$ to label all $n_u$ unlabeled points. Using incremental $k$-centers (Charikar, Chekuri, Feder, & Motwani, 1997) which has provable worst case bound on the distortion, we quantify the error introduced by quantization. Moreover, using regularization we show that the solution is stable, which gives the desired generalization bounds.

The paper is structured as follows. First, we review properties of the regularized HS and discuss its scalability. Second, we propose an online HS and discuss how to implement it efficiently. Third, we prove a performance bound for our online SSL algorithm. Finally, we evaluate our solution on 2 UCI and 2 real-world video datasets and compare it to existing work.

The following notation is used in the paper. The symbols $\mathbf{x}_i$ and $y_i$ refer to the $i$-th training point and its label, respectively. The label $y_i \in \{-1, 1\}$ is binary[2] and never observed for unlabeled data points. The sets of labeled and unlabeled examples are denoted by $l$ and $u$, respectively, and the cardinality of these sets is $n_l = |l|$ and $n_u = |u|$. The number of training examples is $n = n_l + n_u$.

## 2 REGULARIZED HARMONIC SOLUTION

In this section, we build on the harmonic solution (Zhu et al., 2003). Moreover, we show how to regularize it such that it can interpolate between SSL on labeled examples and SSL on all data. A standard approach to semi-supervised learning on graphs is to minimize the quadratic objective function

$$\min_{\ell \in \mathbb{R}^n} \quad \ell^\mathsf{T} L \ell \qquad (1)$$
$$\text{s.t.} \quad \ell_i = y_i \text{ for all } i \in l;$$

where $\ell$ denotes the vector of predictions, $L = D - W$ is the graph Laplacian of the similarity graph, which is represented by a matrix $W$ of weights $w_{ij}$ that encode pairwise similarities, and $D$ is a diagonal matrix whose entries are given by $d_i = \sum_j w_{ij}$. This problem has a closed-form solution which satisfies the *harmonic property* $\ell_i = \frac{1}{d_i} \sum_{j \sim i} w_{ij} \ell_j$, and therefore is commonly known as the *harmonic solution*.

We propose to control the confidence of labeling unlabeled examples by regularizing the Laplacian $L$ as $L + \gamma_g I$, where $\gamma_g$ is a scalar and $I$ is the identity matrix. Similarly to (1), the corresponding problem

$$\min_{\ell \in \mathbb{R}^n} \quad \ell^\mathsf{T}(L + \gamma_g I)\ell \qquad (2)$$
$$\text{s.t.} \quad \ell_i = y_i \text{ for all } i \in l;$$

---

[2]The random walk view of the HS allows for a straightforward generalization to multi-class problems (Balcan, Blum, Choi, Lafferty, Pantano, Rwebangira, & Zhu, 2005).

can be computed in a closed form

$$\ell_u = (L_{uu} + \gamma_g I)^{-1} W_{ul} \ell_l. \quad (3)$$

and we will refer to is as *regularized* HS. It can be also interpreted as a random walk on the graph $W$ with an extra sink. At every step, a walk at node $x_i$ may terminate at the sink with probability $\gamma_g/(d_i + \gamma_g)$ where $d_i$ is the degree of the current node in the walk. Therefore, the scalar $\gamma_g$ essentially controls how the "confidence" $|\ell_i|$ of labeling unlabeled vertices decreases with the number of hops from labeled vertices.

## 3 ALGORITHM

The regularized HS (Section 2) is an offline learning algorithm. This algorithm can be made naïvely online, by taking each new example, connecting it to its neighbors, and recomputing the HS. Unfortunately, this naïve implementation has computational complexity $O(t^3)$ at step $t$, and computation becomes infeasible as more examples are added to the graph.

### 3.1 QUANTIZATION

To address the problem, we use *data quantization* (Gray & Neuhoff, 1998) and substitute the vertices in the graph with a smaller set of $k$ distinct centroids. The resulting $t \times t$ similarity matrix $W$ has many identical rows/columns. We will show that the *exact* HS using $W$ may be reconstructed from a much smaller $k \times k$ matrix $\tilde{W}^q$, where $\tilde{W}^q_{ij}$ contains the similarity between the $i^{th}$ and $j^{th}$ centroids, and a vector $\mathbf{v}$ of length $k$, where $v_i$ denotes to number of points collapsed into the $i^{th}$ centroid. To show this, we introduce the matrix $W^q = V \tilde{W}^q V$ where $V$ is a diagonal matrix containing the counts in $\mathbf{v}$ on the diagonal.

**Proposition 1.** *The harmonic solution (2) using $W$ can be computed compactly as*

$$\ell^q = (L^q_{uu} + \gamma_g V)^{-1} W^q_{ul} \ell_l,$$

*where $L^q$ is the Laplacian of $W^q$.*

**Proof:** Our proof uses the electric circuit interpretation of a random walk (Zhu et al., 2003). More specifically, we show that $W$ and $W^q$ represent identical electric circuits and therefore, their harmonic solutions are the same.

In the electric circuit formulation of $W$, the edges of the graph are resistors with the conductance $w_{ij}$. If two vertices $i$ and $j$ are identical, then by symmetry, the HS must assign the same value to both vertices and we may replace them with a single vertex. Furthermore, they correspond to ends of resistors in parallel. The total conductance of two resistors in parallel

**Inputs:**
 an unlabeled example $\mathbf{x}_t$
 a set of centroids $C_{t-1}$
 vertex multiplicities $\mathbf{v}_{t-1}$

**Algorithm:**
 if ($|C_{t-1}| = k + 1$)
  $R \leftarrow mR$
  greedily repartition $C_{t-1}$ into $C_t$ such that:
   no two vertices in $C_t$ are closer than $R$
   for any $\mathbf{c}_i \in C_{t-1}$ exists $\mathbf{c}_j \in C_t$ such that $d(\mathbf{c}_i, \mathbf{c}_j) < R$
  update $\mathbf{v}_t$ to reflect the new partitioning
 else
  $C_t \leftarrow C_{t-1}$
  $\mathbf{v}_t \leftarrow \mathbf{v}_{t-1}$
 if $\mathbf{x}_t$ is closer than $R$ to any $\mathbf{c}_i \in C_t$
  $\mathbf{v}_t(i) \leftarrow \mathbf{v}_t(i) + 1$
 else
  $\mathbf{v}_t(|C_t| + 1) \leftarrow 1$
  $C_t(|C_t| + 1) \leftarrow \mathbf{x}_t$
 build a similarity matrix $\tilde{W}^q_t$ over the vertices $C_t$
 build a matrix $V_t$ whose diagonal elements are $\mathbf{v}_t$
 $W^q_t = V_t \tilde{W}^q_t V_t$
 compute the Laplacian $L^q$ of the graph $W^q_t$
 infer labels on the graph:
  $\ell^q[t] \leftarrow \arg\min_\ell \ell^\top (L^q + \gamma_g V_t)\ell$
  s.t. $\ell_i = y_i$ for all labeled examples up to time $t$
 make a prediction $\hat{y}_t = \operatorname{sgn}(\ell^q_t[t])$

**Outputs:**
 a prediction $\hat{y}_t$
 a set of centroids $C_t$
 vertex multiplicities $\mathbf{v}_t$

Figure 1: Online quantized harmonic solution at the time step $t$. The mains parameter of the algorithm are the regularizer $\gamma_g$ and the maximum number of centroids $k$.

is equal to the sum of their conductances. Therefore, the two resistors can be replaced by a single resistor with the conductance of the sum. A repetitive application of this rule gives $W^q = V \tilde{W}^q V$, which yields the same HS as $W$. In Section 2, we showed that the regularized HS can be interpreted as having an extra sink in a graph. Therefore, when two vertices $i$ and $j$ are merged, we also need to sum up their sinks. A repetitive application of this rule yields the term $\gamma_g V$ in our closed-form solution. ∎

We note that Proposition 1 may be applied whenever the similarity matrix has identical rows/columns, not just when quantization is applied. However, when the data points are quantized into a fixed number of centroids $k$, it shows that we may compute the HS for the $t^{th}$ point in $O(k^3)$ time. Since the time complexity of computation on the quantized graph is independent of $t$, it gives a suitable algorithm for online learning.

We now describe how to perform incremental quantization with provably nearly-optimal distortion.

## 3.2 INCREMENTAL $k$-CENTERS

We make use of *doubling algorithm* for incremental $k$-center clustering (Charikar et al., 1997) which assigns points to centroids in a near optimal way. In particular, it is a $(1+\epsilon)$-approximation with cost measured by the maximum quantization error over all points. In Section 4.3, we show that under reasonable assumptions, the quantization error goes to zero as the number of centroids increases.

The algorithm of Charikar et al. (1997) maintains a set of centroids $C_t = \{\mathbf{c}_1, \mathbf{c}_2, \dots\}$ such that the distance between any two vertices in $C_t$ is at least $R$ and $|C_t| \leq k$ at the end of each iteration. For each new point $\mathbf{x}_t$, if its distance to some $\mathbf{c}_i \in C_t$ is less than $R$, the point is assigned to $\mathbf{c}_i$. Otherwise, the distance of $\mathbf{x}_t$ to $\mathbf{c}_i \in C_t$ is at least $R$ and $\mathbf{x}_t$ is added to the set of centroids $C_t$. If adding $\mathbf{x}_t$ to $C_t$ results in $|C_t| > k$, the scalar $R$ is doubled and $C_t$ is greedily repartitioned such that no two vertices in $C_t$ are closer than $R$. The doubling of $R$ also ensures that $|C_t| < k$.

Pseudocode of our algorithm is given in Figure 1. We make a small modification to the original quantization algorithm in that, instead of doubling $R$, we multiply it with some $m > 1$. This still yields a $(1+\epsilon)$-approximation algorithm as it still obeys the invariants given in Lemma 3.4 in Charikar et al. (1997). We also maintain a vector of multiplicities $\mathbf{v}$ which contains the number of vertices that each centroid represents. At each time step, the HS is calculated using the updated quantized graph, and a prediction is made.

The incremental $k$-centers algorithm also has the advantage that it provides a variable $R$ which may be used to bound the maximum quantization error. In particular, at any point in time $t$, the distance of any example from its centroid is at most $Rm/(m-1)$. To see this, consider the following. As the new data arrive we keep increasing $R$ by multiplying it by some $m > 1$. But for any point at any time, the centroid assigned to a vertex is at most $R$ apart from the previously assigned centroid, which is at most $R/m$ apart from the centroid assigned before, etc. Summing up, at any time, any point is at most

$$R + \frac{R}{m} + \frac{R}{m^2} + \cdots = R\left(1 + \frac{1}{m} + \frac{1}{m^2} + \cdots\right) = \frac{Rm}{m-1}$$

apart from its assigned centroid, where $R$ is the most recent one.

## 4 THEORETICAL ANALYSIS

In the rest of this paper, $W$ denotes full data similarity matrix, $W_t^o$ its observed portion up to time $t$ and $W_t^q$ the quantized version of $W_t^o$. For simplicity, we do not consider the compact version of quantized matrix. In other words, $W_t^q$ is $t \times t$ matrix with at most $k$ distinct rows/columns. The Laplacians and regularized Laplacians of these matrices are denoted as $L, L^o, L^q$ and $K, K^o, K^q$ respectively. Similarly, we use $\ell^*, \ell^o[t]$, and $\ell^q[t]$ to refer to the harmonic solutions on $W$, $W_t^o$, and $W_t^q$ respectively. Finally, $\ell_t^*$, $\ell_t^o[t]$, and $\ell_t^q[t]$ refer to the predicted label of the example $\mathbf{x}_t$.

In this section, we use a stability argument to bound quality of the predictions. We note that the derived bounds are not tight. Our online learner (Figure 1) solves an online regression problem. As a result, it should ideally minimize the error of the form $\sum_t (\ell_t^q[t] - y_t)^2$, where $\ell_t^q[t]$ is the prediction at the time step $t$ (again, time is denoted in the square brackets). In the following proposition we decompose this error into three terms. The first term (4) corresponds to the generalization error of the HS and is bounded by the algorithm stability argument. The second term (5) appears in our online setting because the similarity graph is only partially revealed. Finally, the third term (6) quantifies the error introduced due to quantization of the similarity matrix.

**Proposition 2.** *Let $\ell_t^q[t]$, $\ell_t^o[t]$, $\ell_t^*$ be the predictions as defined above and let $y_t$ be the true labels. Then the error of our predictions $\ell_t^q[t]$ is bounded as*

$$\frac{1}{n}\sum_{t=1}^n (\ell_t^q[t] - y_t)^2 \leq \frac{9}{2n}\sum_{t=1}^n (\ell_t^* - y_t)^2 \quad (4)$$

$$+ \frac{9}{2n}\sum_{t=1}^n (\ell_t^o[t] - \ell_t^*)^2 \quad (5)$$

$$+ \frac{9}{2n}\sum_{t=1}^n (\ell_t^q[t] - \ell_t^o[t])^2. \quad (6)$$

**Proof:** Our bound follows from the inequality

$$(a-b)^2 \leq \frac{9}{2}\left[(a-c)^2 + (c-d)^2 + (d-b)^2\right],$$

which holds for $a, b, c, d \in [-1, 1]$. ∎

We continue by bounding all the three sums in the Proposition 2. These sums can be bounded if the constraints $\ell_i = y_i$ are enforced in a soft manner (Cortes, Mohri, Pechyony, & Rastogi, 2008). One way of achieving this is by solving a related problem

$$\min_{\ell \in \mathbb{R}^n} (\ell - \mathbf{y})^\intercal C(\ell - \mathbf{y}) + \ell^\intercal K\ell, \quad (7)$$

where $K = L + \gamma_g I$ is the regularized Laplacian of the similarity graph, $C$ is a diagonal matrix such that $C_{ii} = c_l$ for all labeled examples, and $C_{ii} = c_u$ otherwise, and $\mathbf{y}$ is a vector of pseudo-targets such that $y_i$ is the label of the $i$-th example when the example is labeled, and $y_i = 0$ otherwise.

## 4.1 BOUNDING $\frac{1}{n}\sum_{t=1}^{n}(\ell_t^* - y_t)^2$

The following proposition bounds the generalization error of the solution to the problem (7). We then use it to bound the HS part (4) of Proposition 2.

**Proposition 3.** *Let $\ell^*$ be a solution to the problem (7), where all labeled examples $l$ are selected i.i.d. If we assume that $c_l = 1$ and $c_l \gg c_u$, then the inequality*

$$R(\ell^*) \leq \widehat{R}(\ell^*) + \underbrace{\beta + \sqrt{\frac{2\ln(2/\delta)}{n_l}}(n_l\beta + 4)}_{\text{transductive error } \Delta_T(\beta, n_l, \delta)}$$

$$\beta \leq 2\left[\frac{\sqrt{2}}{\gamma_g + 1} + \sqrt{2n_l}\frac{1-\sqrt{c_u}}{\sqrt{c_u}}\frac{\lambda_M(L)+\gamma_g}{\gamma_g^2 + 1}\right]$$

*holds with probability $1 - \delta$, where*

$$R(\ell^*) = \frac{1}{n}\sum_{t}(\ell_t^* - y_t)^2 \text{ and } \widehat{R}(\ell^*) = \frac{1}{n_l}\sum_{t \in l}(\ell_t^* - y_t)^2$$

*are risk terms for all and labeled vertices, respectively, and $\beta$ is the stability coefficient of the solution $\ell^*$.*

The proof can be found in Kveton et al. (2010b). Proposition 3 shows that when $\Delta_T(\beta, n_l, \delta) = o(1)$, the true risk is not much different from the empirical risk on the labeled points which bounds the generalization error. This occurs when $\beta = o(n_l^{-1/2})$, which corresponds to setting $\gamma_g = \Omega(n_l^{1+\alpha})$ for any $\alpha > 0$.

## 4.2 BOUNDING $\frac{1}{n}\sum_{t=1}^{n}(\ell_t^o[t] - \ell_t^*)^2$

In the following, we will bound the difference between the online and offline HS and use it to bound (5) of the Proposition 2. The idea is that when Laplacians $L$ and $L^o$ are regularized enough by $\gamma_g$, resulting harmonic solutions are close to zero a therefore close to each other. We first show that any regularized HS can be bounded as follows:

**Lemma 1.** *Let $\ell$ be a regularized harmonic solution, i.e. $\ell = (C^{-1}K + I)^{-1}\mathbf{y}$ where $K = L + \gamma_g I$. Then*

$$\|\ell\|_2 \leq \frac{\sqrt{n_l}}{\gamma_g + 1}.$$

**Proof:** If $A \in \mathbb{R}^{n \times n}$ is a symmetric matrix and $\lambda_m(A)$ and $\lambda_M(A)$ are its smallest and largest eigenvalues, then for any $\mathbf{v} \in \mathbb{R}^{n \times 1}$, $\lambda_m(A)\|\mathbf{v}\|_2 \leq \|A\mathbf{v}\|_2 \leq \lambda_M(A)\|\mathbf{v}\|_2$. Then

$$\|\ell\|_2 \leq \frac{\|\mathbf{y}\|_2}{\lambda_m(C^{-1}K + I)} = \frac{\|\mathbf{y}\|_2}{\frac{\lambda_m(K)}{\lambda_M(C)} + 1} \leq \frac{\sqrt{n_l}}{\gamma_g + 1}. \quad \blacksquare$$

The straightforward implication of Lemma 1 is that any 2 regularized harmonic solutions can be bounded as in the following proposition:

**Proposition 4.** *Let $\ell^o[t]$ be the predictions of the online HS, and $\ell^*$ be the predictions of the offline HS. Then*

$$\frac{1}{n}\sum_{t=1}^{n}(\ell_t^o[t] - \ell^*[t])^2 \leq \frac{4n_l}{(\gamma_g + 1)^2}. \quad (8)$$

**Proof:** We use the fact that $\|\cdot\|_2$ is an upper bound on $\|\cdot\|_\infty$. Therefore, for any $t$

$$(\ell_t^o[t] - \ell_t^*)^2 \leq \|\ell^o[t] - \ell^*\|_\infty^2 \leq \|\ell^o[t] - \ell^*\|_2^2$$
$$\leq \left(\frac{2\sqrt{n_l}}{\gamma_g + 1}\right)^2,$$

where in the last step we used Lemma 1 twice. By summing over $n$ and dividing by $n$ we get (8). $\blacksquare$

From Proposition 4 we see that we can achieve convergence of the term (5) at the rate of $O(n^{-1/2})$ with $\gamma_g = \Omega(n^{1/4})$.

## 4.3 BOUNDING $\frac{1}{n}\sum_{t=1}^{n}(\ell_t^q[t] - \ell_t^o[t])^2$

In this section, we show in Proposition 5 a way to bound the error for the HS between the full and quantized graph, and then use it to bound the difference between the *online* and *online quantized* HS in (6). Let us consider the perturbed version of the problem (7), where we replace the regularized Laplacian $K^o$ with $K^q$; i.e., $K^q$ corresponds to the regularized Laplacian of the quantized graph. Let $\ell^o$ and $\ell^q$ minimize (7) and its perturbed version respectively. Their closed-form solutions are given by $\ell^o = (C^{-1}K^o + I)^{-1}\mathbf{y}$ and $\ell^q = (C^{-1}K^q + I)^{-1}\mathbf{y}$ respectively. We now follow the derivation of Cortes et al. (2008) that derives stability coefficient of unconstrained regularization algorithms. Instead of considering perturbation on $C$, we consider the perturbation on $K^o$. Our goal is to derive a bound on a difference in HS when we use $K^q$ instead of $K^o$.

**Lemma 2.** *Let $\ell^o$ and $\ell^q$ minimize (7) and its perturbed version respectively. Then*

$$\|\ell^q - \ell^o\|_2 \leq \frac{\sqrt{n_l}}{c_u \gamma_g^2}\|K^q - K^o\|_F.$$

**Proof:** Let $Z^q = C^{-1}K^q + I$ and $Z^o = C^{-1}K^o + I$. By definition

$$\ell^q - \ell^o = (Z^q)^{-1}\mathbf{y} - (Z^o)^{-1}\mathbf{y} = (Z^q Z^o)^{-1}(Z^o - Z^q)\mathbf{y}$$
$$= (Z^q Z^o)^{-1}C^{-1}(K^o - K^q)\mathbf{y}.$$

Using the eigenvalue inequalities from the proof of Lemma 1 we get

$$\|\ell^q - \ell^o\|_2 \leq \frac{\lambda_M(C^{-1})\|(K^q - K^o)\mathbf{y}\|_2}{\lambda_m(Z^q)\lambda_m(Z^o)}. \quad (9)$$

By the compatibility of $||\cdot||_F$ and $||\cdot||_2$ and since $\mathbf{y}$ is zero on unlabeled points, we have

$$\|(K^{\mathrm{q}}-K^{\mathrm{o}})\mathbf{y}\|_2 \leq \|K^{\mathrm{q}}-K^{\mathrm{o}}\|_F \cdot \|\mathbf{y}\|_2 \leq \sqrt{n_l}\|K^{\mathrm{q}}-K^{\mathrm{o}}\|_F.$$

Furthermore,

$$\lambda_m(Z^{\mathrm{o}}) \geq \frac{\lambda_m(K^{\mathrm{o}})}{\lambda_M(C)}+1 \geq \gamma_g \quad \text{and} \quad \lambda_M(C^{-1}) \leq c_u^{-1},$$

where $c_u$ is a small constant as defined in (7). By plugging these inequalities into (9) we get the desired bound. ∎

**Proposition 5.** *Let $\ell_t^{\mathrm{q}}[t]$ be the predictions of the online harmonic solution on the quantized graph at the time step $t$, $\ell_t^{\mathrm{o}}[t]$ be predictions of the online harmonic solution at the time step $t$. Then*

$$\frac{1}{n}\sum_{t=1}^n (\ell_t^{\mathrm{q}}[t] - \ell_t^{\mathrm{o}}[t])^2 \leq \frac{n_l}{c_u^2 \gamma_g^4}\|L^{\mathrm{q}} - L^{\mathrm{o}}\|_F^2. \qquad (10)$$

**Proof:** Similarly as in Proposition 4, we get

$$(\ell_t^{\mathrm{q}}[t] - \ell_t^{\mathrm{o}}[t])^2 \leq \|\ell^{\mathrm{q}}[t] - \ell^{\mathrm{o}}\|_\infty^2 \leq \|\ell^{\mathrm{q}}[t] - \ell^{\mathrm{o}}\|_2^2$$
$$\leq \left(\frac{\sqrt{n_l}}{c_u \gamma_g^2}\|K^{\mathrm{q}} - K^{\mathrm{o}}\|_F\right)^2,$$

where we used (9) the last step. We also note that

$$K^{\mathrm{q}} - K^{\mathrm{o}} = L^{\mathrm{q}} + \gamma_g I - (L^{\mathrm{o}} + \gamma_g I) = L^{\mathrm{q}} - L^{\mathrm{o}},$$

which gives us $(\ell_t^{\mathrm{q}}[t] - \ell_t^{\mathrm{o}}[t])^2 \leq \|L^{\mathrm{q}} - L^{\mathrm{o}}\|_F^2 \cdot n_l/(c_u^2 \gamma_g^4)$. By summing over $n$ and dividing by $n$ we get (10). ∎

If $\|L^{\mathrm{q}} - L^{\mathrm{o}}\|_F^2 = O(1)$, the left-hand side of (10) converges to zero at the rate of $O(n^{-1/2})$ with $\gamma_g = \Omega(n^{1/8})$. We show this condition is achievable whenever the Laplacian is scaled appropriately. Specifically, we demonstrate that normalized Laplacian achieves this bound when the quantization is performed using incremental $k$-center clustering in Section 3.1, and when the weight function obeys a Lipschitz condition (e.g. the Gaussian kernel). We also show that this error goes to zero as the number of center points $k$ goes to infinity. This result is directly applicable to unnormalized Laplacian used in previous sections, with details being omitted due to space limitation.

Suppose the data $\{\mathbf{x}_i\}_{i=1,...,n}$ lie on a smooth $d$-dimensional compact manifold $\mathcal{M}$ with boundary of bounded geometry embedded in $\mathbb{R}^b$. We first demonstrate that the distortion introduced by quantization is small, and then show that small distortion gives small error in the Frobenius norm.

**Proposition 6.** *Using incremental $k$-center clustering for quantization has maximum distortion $Rm/(m-1) = \max_{i=1,...,n} \|\mathbf{x}_i - \mathbf{c}\|_2 = O(k^{-1/d})$, where $\mathbf{c}$ is the closest centroid to $\mathbf{x}_i$.*

**Proof:** Consider a sphere packing with $k$ centers contained in $\mathcal{M}$ and each with radius $r$. Since the manifold is compact and the boundary has bounded geometry, it has finite volume $V$ and finite surface area $A$. The maximum volume that the packing can occupy obeys the inequality $kc_d r^d \leq V + Ac_\mathcal{M} r$ for some constants $c_d, c_\mathcal{M}$ that only depend on the dimension and the manifold. Trivially, if $k$ is sufficiently large, then $r < 1$, and we have an upper bound $r < ((V + Ac_\mathcal{M})/(kc_d))^{1/d} = O(k^{-1/d})$. An $r$-packing is a $2r$-covering, so we have an upper bound on the distortion of the optimal $k$-centers solution. Since the incremental $k$-centers algorithm is a $(1+\epsilon)$-approximation algorithm (Charikar et al., 1997), it follows that the maximum distortion returned by the algorithm is $Rm/(m-1) = 2(1+\epsilon)O(k^{-1/d})$. ∎

We now show that with appropriate normalization, the error $\|L^{\mathrm{q}} - L^{\mathrm{o}}\|_F^2 = O(k^{-2/d})$. If $L^{\mathrm{q}}$ and $L^{\mathrm{o}}$ are normalized Laplacians, then this bound holds if the underlying density is bounded away from 0. Note that since we use the Gaussian kernel, the Lipschitz condition is satisfied.

**Proposition 7.** *Let $W_{ij}^{\mathrm{o}}$ be a weight matrix constructed from $\{x_i\}_{i=1,...,n}$ and a bounded, Lipschitz function $\omega(\cdot,\cdot)$ with Lipschitz constant $M$. Let $D^{\mathrm{o}}$ be the corresponding degree matrix and $L_{ij}^{\mathrm{o}} = (D_{ij}^{\mathrm{o}} - W_{ij}^{\mathrm{o}})/c_{ij}^{\mathrm{o}}$ be the normalized Laplacian. Suppose $c_{ij}^{\mathrm{o}} = \sqrt{D_{ii}^{\mathrm{o}} D_{jj}^{\mathrm{o}}} > c_{min} n$ for some constant $c_{min} > 0$ that does not depend on $k$. Likewise define $W^{\mathrm{q}}, L^{\mathrm{q}}, D^{\mathrm{q}}$ on the quantized points. Let the maximum distortion be $Rm/(m-1) = O(k^{-1/d})$. Then $\|L^{\mathrm{q}} - L^{\mathrm{o}}\|_F^2 = O(k^{-2/d})$.*

**Proof:** Since $\omega$ is Lipschitz, we have that $|W_{ij}^{\mathrm{q}} - W_{ij}^{\mathrm{o}}| < 2MRm/(m-1)$ and $|c_{ij}^{\mathrm{q}} - c_{ij}^{\mathrm{o}}| < 2nMRm/(m-1)$. The error of a single off-diagonal entry of the Laplacian matrix is

$$\begin{aligned}
L_{ij}^{\mathrm{q}} - L_{ij}^{\mathrm{o}} &= \frac{W_{ij}^{\mathrm{q}}}{c_{ij}^{\mathrm{q}}} - \frac{W_{ij}^{\mathrm{o}}}{c_{ij}^{\mathrm{o}}} \\
&\leq \frac{W_{ij}^{\mathrm{q}} - W_{ij}^{\mathrm{o}}}{c_{ij}^{\mathrm{q}}} + \frac{W_{ij}^{\mathrm{q}}(c_{ij}^{\mathrm{q}} - c_{ij}^{\mathrm{o}})}{c_{ij}^{\mathrm{o}} c_{ij}^{\mathrm{q}}} \\
&\leq \frac{4MRm}{(m-1)c_{min}n} + \frac{4M(nMRm)}{((m-1)c_{min}n)^2} \\
&= O\left(\frac{R}{n}\right).
\end{aligned}$$

The error on the diagonal entries is 0 since the diagonals of $L^{\mathrm{q}}$ and $L^{\mathrm{o}}$ contain all 1. Thus $\|L^{\mathrm{q}} - L^{\mathrm{o}}\|_F^2 \leq n^2 O(R^2/n^2) = O(k^{-2/d})$. ∎

Here we showed the asymptotic behavior $\|L^{\mathrm{q}} - L^{\mathrm{o}}\|_F$ in term of the number of vertices used in the quantized graph. In Section 5, we empirically show that $\|L^{\mathrm{q}} - L^{\mathrm{o}}\|_F$ vanishes quickly as the number of vertices increases (Figure 3). Moreover, with fixed number of vertices, $\|L^q - L^o\|_F$ quickly flattens out even when the data size (time) keeps increasing (Figure 2).

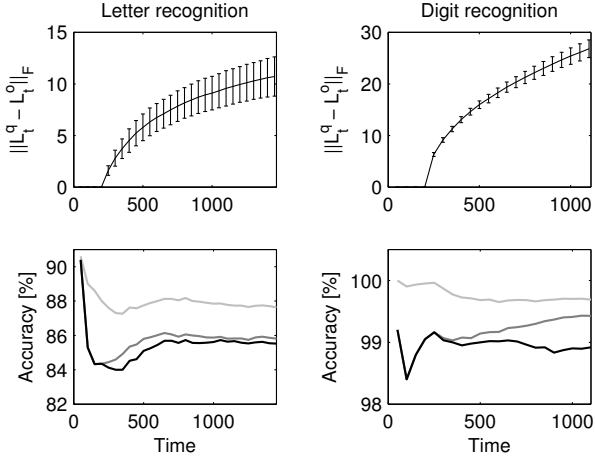
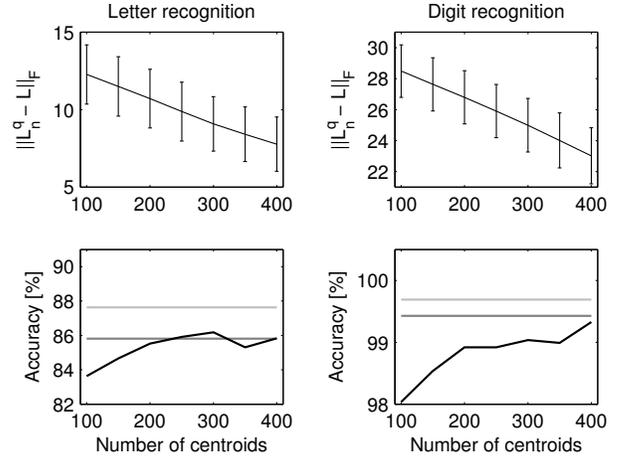

Figure 2: The upper plots show the difference between the normalized Laplacian $L_t^o$ and its approximation $L_t^q$ at time $t$. The bottom plots show the cumulative accuracy of the harmonic solutions on $W$ (light gray lines), $W_t^o$ (dark gray lines), and $W_t^q$ (black lines) for various times $t$.

### 4.4 DISCUSSION

Our goal in this section is to show how much of regularization $\gamma_g$ is needed for error of our predictions to reasonably decrease over time. We point out that in Proposition 3 the lower bound for $\gamma_g$ for reasonable convergence is a function of $n_l$ labeled examples. On the other hand, in Propositions 4 and 5 those lower bounds are the functions of all $n$ examples.

In particular, Proposition 3 requires $\gamma_g = \Omega(n_l^{1+\alpha})$, $\alpha > 0$ for the true risk not to be much different from the empirical risk on the labeled points. Next, Propositions 4 and 5 require $\gamma_g = \Omega(n^{1/4})$ and $\gamma_g = \Omega(n^{1/8})$ respectively for the terms (5) and (6) to be $O(n^{-1/2})$.

For many applications of online SSL, a small set of $n_l$ labeled example is given in advance, the rest of the examples are unlabeled. That means we usually expect $n \gg n_l$. Therefore, if we regard $n_l$ as a constant, we need to regularize as much as $\gamma_g = \Omega(n^{1/4})$. For such a setting of $\gamma_g$ we have that for $n$ approaching infinity the error of our predictions is getting close to the empirical risk on labeled examples with the rate of $O(n^{-1/2})$.

## 5 EXPERIMENTS

The experimental section is divided into two parts. In the first part, we evaluate our online learner (Figure 1) on UCI ML repository datasets (Asuncion & Newman,

Figure 3: The upper plots show the difference between the normalized Laplacian $L$ and its approximation $L_n^q$. The difference is plotted as a function of the number of centroids $k$. The bottom plots compare the cumulative accuracy of the harmonic solutions up to time $n$ on $W$ (light gray lines), $W_t^o$ (dark gray lines), and $W_t^q$ (black lines).

2007). In the second part, we apply our learner to solve two face recognition problems. In all experiments, the multiplicative parameter $m$ of the $k$-centers algorithm is set to $1.5$.

### 5.1 UCI ML REPOSITORY DATASETS

In the first experiment, we study the online quantization error $\|L_t^q - L_t^o\|_F$ and its relation to the HS on the quantized graphs $W_t^q$. This experiment is performed on two datasets from the UCI ML repository: letter and optical digit recognition. The datasets are converted into a set of binary problems, where each class is discriminated against every other class. The similarity weights computed as $w_{ij} = \exp[-\|\mathbf{x}_i - \mathbf{x}_j\|_2^2 / (2p\sigma^2)]$, where $p$ is the number of features and $\sigma$ denotes the mean of their standard deviations. Our results are averaged over 10 problems from each dataset, and shown in Figures 2 and 3.

In Figure 2, we fix the number of centroids at $k = 200$ and study how the quality of our solution changes with the learning time $t$. Two trends are apparent. First, as time $t$ increases, the error $\|L_t^q - L_t^o\|_F$ slowly levels off. Second, the accuracy of the harmonic solutions on $W_t^q$ changes little with $t$. These trends indicate that a fixed number of centroids $k$ may be sufficient for quantizing similarity graphs that grow with time. In Figure 3, we fix the learning time at $t = n$ and vary the number of centroids $k$. Note that as $k$ increases, the quantization error decreases and the quality of the solutions on $W_t^q$ improves. This trend is consistent with the theoretical results in our paper.

## 5.2 FACE RECOGNITION

In the second experiment, we evaluate our learner on 2 face recognition datasets: office space and environment adaptation. The environment adaptation dataset consists of faces of a single person, which are captured at various locations, such as a cubicle, a conference room, and the corner with a couch (Figure 4). The first four faces in the cubicle are labeled and we want to learn a face recognizer for all locations. To test the sensitivity of the recognizer to outliers, we appended the dataset by random faces. The office space dataset (Figure 4) is multi-class, and involves 8 people who walk in front of a camera and make funny faces. When a person shows up on the camera for the first time, we label four faces of the person. Our goal is to learn good face recognizers for all 8 people.

The similarity of faces $\mathbf{x}_i$ and $\mathbf{x}_j$ is computed as $w_{ij} = \exp\left[-d(\mathbf{x}_i, \mathbf{x}_j)^2 / 2\sigma^2\right]$, where $\sigma$ is a heat parameter, which is set to $\sigma = 0.025$, and $d(\mathbf{x}_i, \mathbf{x}_j)$ is the distance of the faces in the feature space. To make the graph $W$ sparse, we treat it as an $\varepsilon$-neighborhood graph and set $w_{ij}$ to 0 when $w_{ij} < \varepsilon$. The scalar $\varepsilon$ is set as $\varepsilon = 0.1\gamma_g$. As a result, the lower the regularization parameter $\gamma_g$, the higher the number of edges in the graph $W$ and our learner extrapolates to more unlabeled examples. If an example is disconnected from the rest of the graph $W$, we treat it as an outlier, and neither predict the label of the example nor use it to update the quantized graph. This setup makes our algorithm robust to outliers, and allows for controlling its precision and recall by a single parameter $\gamma_g$. In the rest of the section, the number of centroids $k$ is fixed at 500. A more detailed description of the experimental setup is in Kveton et al. (2010a).

In Figure 5, we compare our online algorithm to online semi-supervised boosting (Grabner et al., 2008) and a nearest-neighbor (NN) classifier, which is trained on all labeled faces. The algorithm of Grabner *et al.* (2008) is modified to allow for a fair comparison to our method. First, all weak learners have the nearest-neighbor form $h_i(\mathbf{x}_t) = \mathbb{1}\{w_{it} \geq \varepsilon\}$, where $\varepsilon$ is the radius of the neighborhood. Second, outliers are modeled implicitly. The new algorithm learns a regressor $H(\mathbf{x}_t) = \sum_i \alpha_i h_i(\mathbf{x}_t)$, which yields $H(\mathbf{x}_t) = 0$ for outliers and $H(\mathbf{x}_t) > 0$ when the detected face is recognized.

Figure 5a clearly shows that our learner is better than the nearest-neighbor classifier. Furthermore, note that online semi-supervised boosting yields as good results as our method when given a good set of weak learners. However, future data are rarely known in advance, and when the weak learners are chosen using only a part of the dataset, the quality of the boosted results degrades significantly (Figure 5a). In comparison, our algorithm constantly adapts its representation of the world. How to incorporate a similar adaptation step in online semi-supervised boosting is not obvious.

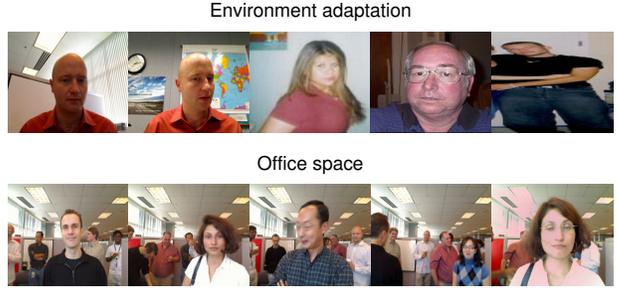

Figure 4: Snapshots from the environment adaptation and office space datasets.

In Figure 5b, we evaluate our learner on an 8-class face recognition problem. Despite the fact that only 4 faces of each person are labeled, we can identify people with 95 percent precision and 90 percent recall. In general, our precision is 10 percent higher than the precision of the NN classifier at the same recall level.

## 6 CONCLUSION

We have presented a fast approximate algorithm for online semi-supervised learning. Our algorithm leverages the incremental $k$-center quantization method to group neighboring points to produce a set of reduced representative points, and to construct an approximate similarity graph for a harmonic solution. This algorithm significantly reduces the expense of the matrix computation in the harmonic solution, while retaining good control on the classification accuracy. Our evaluation shows that a significant speedup for semi-supervised learning can be achieved with little degradation in classification accuracy. Our theoretical analysis reveals that the degradation in classification accuracy is closely related to the amount of data quantization: one can make the degradation smaller by reducing the amount of quantization.

In future work, we would like to develop other online data reduction methods (e.g., data squashing and condensation methods) for preprocessing, and extend our bounds to these methods. We also plan to apply our online learner to other domains, where streams of unlabeled data are readily available, such as object recognition and augmented reality.

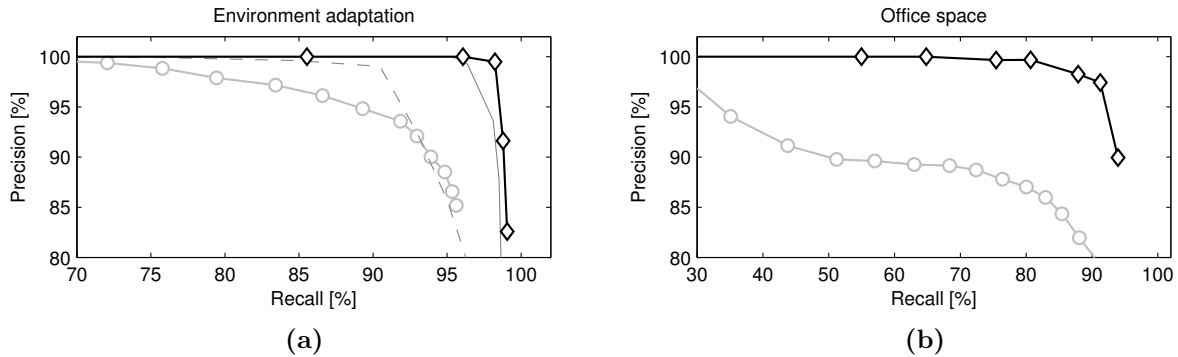

Figure 5: Comparison of 3 face recognizers on 2 face recognition datasets. The recognizers are trained by a NN classifier (gray lines with circles), online semi-supervised boosting (thin gray lines), and our online learner (black lines with diamonds). The plots are generated by varying the parameters $\varepsilon$ and $\gamma_g$. From left to right, the points on the plots correspond to decreasing values of the parameters. Online semi-supervised boosting is performed on 500 weak NN learners, which are sampled at random from the whole environment adaptation dataset (solid line), and its first and last quarters (dashed line).